# Cloud-based Automatic Speech Recognition Systems for Southeast Asian Languages

Lei Wang, Rong Tong, Cheung-Chi Leung, Sunil Sivadas, Chongjia Ni, and Bin Ma
Human Language Department
Institute for Infocomm Research, Singapore
{wangl, tongrong, ccleung, sivadass, nicj, mabin}@i2r.a-star.edu.sg

*Abstract*—This paper provides an overall introduction of our Automatic Speech Recognition (ASR) systems for Southeast Asian languages. As not much existing work has been carried out on such languages, a few difficulties should be addressed before building the systems: limitation on speech and text resources, lack of linguistic knowledge, etc. This work takes Bahasa Indonesia and Thai as examples to illustrate the strategies of collecting various resources required for building ASR systems.

## I. INTRODUCTION

Automatic Speech recognition (ASR) technology has been successfully commercialized and adopted by millions of smart device end users during the past years. Most of the ASR applications are cloud-based due to the rapid development of Internet and smart devices. For instances, Apple's Siri, voice search and dictation are frequently used on smart phones and tablets. In addition, ASR serves as one of the warming care technologies and it can benefit people with hearing loss by converting human speech to text.

Most of the state-of-the-art ASR systems focus on popular languages, e.g. English and Mandarin, due to the high demand of such languages. Effort has also been spent to collect enough resources to carry out the work. Figure 1 shows the system structure of a typical ASR system. For popular languages, large amounts of transcribed speech data and text data are available for building the Acoustic Models (AMs) and Language Models (LMs). To generate pronunciation dictionaries, various types of acoustic units and pronunciation rules have been examined by linguists. For example, pronunciation of Mandarin words can be represented using phoneme-based units [1] or initial-final units [2]. The sufficient speech resources can be used to build a recognizer with relatively high accuracy.

Different from the popular languages, ASR of Southeast Asian languages did not attract sufficient attention in the speech community. However, we observed a trend of increment in market demand of such languages due to the large population of the speakers. For examples, there were 77 million native Malay speakers (including Malaysian and Indonesian), and 56 million native Thai speakers in 2007 [3]. In such emerging markets, the young generations are optimistic and they are willing to spend on the smart devices and applications. Motivated by the demand, this work aims to develop large vocabulary continuous speech recognition systems for Southeast Asian languages.

Existing research work on Southeast Asian languages was usually with specific applications such as keyword spotting [4-6]. For example, Vietnamese [4] and Tamil [5-6] were adopted as the low-resource languages by the NIST Open Keyword Search Evaluation 2013 and 2014, respectively. However, ASR systems of these languages have not been widely studied except a few preliminary attempts.

Among the existing ASR work of Southeast Asian languages, Sinaporn *et al.* [7] developed a Thai speech recognizer using tens of hours' data and the recognizer was applied in a Thai-to-English speech-to-speech translation system. To recognize Vietnamese, an ASR system was reported in [8] and the AM was trained using Vietnamese portion of Globalphone corpus from multilingual phone inventory. A Myanmar ASR system was also reported in [9] and the work compared the effectiveness of different language modeling and acoustic modeling techniques.

Besides the above mentioned traditional approaches of building AMs, the state-of-the-art deep learning techniques have also been applied. Researchers in [10] built a speech recognition system for Tamil language which is predominantly spoken in the South of India, Sri Lanka, Singapore and Malaysia. A Deep Neural Network (DNN) based AM was trained using 150-hour Tamil speech data collected from about 500 native Tamil speakers. In another work, a deep maxout network structure [11] was examined to improve the performance of Tagalog speech recognition.

Researchers also attempted to address a few other issues in ASR of Southeast Asian languages. To build a Malay ASR system, effort has been made to examine a rule-based grapheme-to-phoneme (G2P) algorithm [12] to automatically generate the pronunciation dictionary. Experimental results on a 70-hour Malay corpus showed that the ASR performance using the G2P-based dictionary was comparable to the manually verified dictionary. In another work of Indonesian ASR [13], researchers applied maximum *a posteriori* (MAP) to adapt the read speech AM towards spontaneous domain by experimenting on 152-hour read speech data and 44.5-hour spontaneous data.

Ethnic and accent diversity is another issue observed in Southeast Asian languages. Researchers in [14] presented a Malay speech recognition system and it was evaluated using test data collected from speakers with various accents. Sakriani *et al.* carried out ASR work on four Indonesian ethnic languages [15]. In [16], researchers applied low-resource acoustic modeling strategies on Vietnamese and Khmer.

Several common practical challenges observed in the existing work include: i) insufficient manually transcribed and recorded speech data; ii) insufficient text data; and iii) lack of knowledge in pronunciation rules. To overcome the challenges, this work proposes general strategies and guidelines to collect speech and text data of Southeast Asian languages. Moreover, we also propose to use International Phonetic Alphabet (IPA)-based dictionaries hence the acoustic models are defined as IPA elements.

The above mentioned strategies were applied on different Southeast Asian ASR systems including Malay, Tamil, Bahasa Indonesia, Thai, Vietnamese and Cantonese. The experimental results showed that reasonable ASR performance could be achieved on open test sets. This work uses Bahasa Indonesia and Thai as case studies.

The remaining of the paper is organized as below: Section II describes the guidelines to collect speech corpora, followed by introducing pronunciation dictionaries and strategies in text data collection in Section III. Section IV presents experimental setup and results, and Section V describes the cloud-based application of ASR. Eventually, we conclude in Section VI.

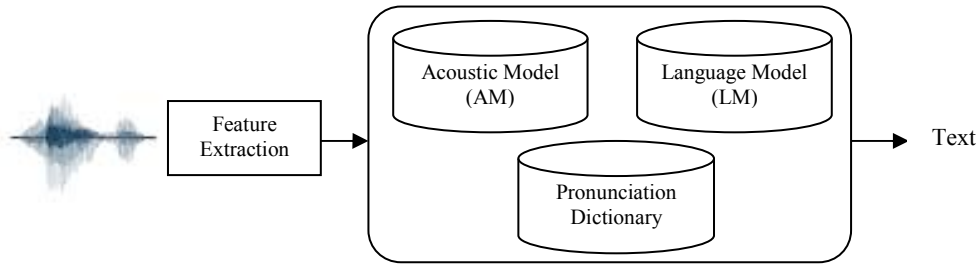

Figure 1. A typical Speech Recognition System consists of 4 major models: A robust front-end algorithm for Feature Extraction; Large amount of transcribed speech data is required to train AM; Large amount of text data is required to train LM; Pronunciation rules are required to generate a pronunciation dictionary.

## II. COLLECTION OF SPEECH CORPORA

To build a general-purpose ASR system, speech samples are better to have a wide coverage of speakers, vocabulary, recording devices (or channels), environments, etc. A wide coverage of speech samples has been proved to improve the robustness of ASR systems.

### A. Speaker Coverage

The physical or behavioral attributes are usually distinctive and measurable characteristics to differentiate human individuals. Speech samples collected from a small group of speakers may result the ASR system to have bias to a particular group of users. To capture the variability of speech samples, the below criteria are proposed to guide the data collection:

i) A sufficient number of speakers should be involved in the study. Depending on the scale of the system, usually hundreds to thousands of speakers are recruited;
ii) The speakers should be gender balanced;
iii) The speakers should be from a wide age range, e.g. balanced from 16 to 60-year old;
iv) The speakers should come from different regions of the country so that speakers with different accents [14-16] can be recruited;
v) Native speakers are preferred.

Due to multiracialism and multiculturalism in Southeast Asian countries, many people are able to speak more than one language. The speech samples should be collected from speakers whose mother language is the target language.

### B. Vocabulary Size and Content

To record the voice from recruited speakers, prompts can be designed and provided to them hence they can follow the scripts to speak. The scripts usually contain a set of literally correct sentences and the sentences should be phonetically rich. They can be selected from different domains or media: news, chats, Twitter, etc. Each sentence may contain less than 20 words and a unique sentence does not appear more than 3 times across the entire corpus. The entire scripts should cover as many words as possible.

### C. Recording Channels and Devices

Recording channels are usually classified as narrowband (e.g. telephony channel of 8KHz sampling rate) or wideband (e.g. mobile devices of 16KHz sampling rate). To increase the capability of the ASR systems, speech samples can be recorded using various devices. For example, the 16KHz speech samples can be recorded using mainstream mobile devices running iOS, Android, Windows and so on.

### D. Environments

The recording environments are generally categorized into 2 groups: quiet (or clean) and noisy. The quiet environments include office, home, and other indoor environment without echo; while noisy environments include streets, shopping malls, in-car, restaurants, etc. A mix of samples recorded from different environments will improve the robustness of the ASR system. However, the cost of recording in noisy environments is usually higher than quiet environments. This work focuses on data collection in quiet environments.

By following the above guidelines, the collected speech samples together with the scripts will be used to train the AM of the ASR system.

## III. PRONUNCIATION DICTIONARY AND TEXT DATA COLLECTION

### A. Pronunciation Dictionary

Pronunciation dictionary is one of the key components in the ASR system. Existing studies [1-2] proposed different types of phoneme inventory for the same language to achieve either lower Word Error Rate (WER) or higher real time factor. This section briefly describes the dictionaries we adopted for Bahasa Indonesia and Thai languages.

Bahasa Indonesia is written using the Latin script and its alphabets are based on the 26 Latin letters. The basic acoustic units of the language are phonemes, and one straightforward way of defining phoneme inventory is to follow its IPA definition. There are 35 such phonemes including vowels, semi-vowels, plosives, nasals, fricatives and so on.

Thai belongs to the Tai group of the Tai-Kadai language family and it is a tonal language. The definition of its phoneme inventory also follows IPA, and it consists of 42 phonemes including consonants, vowels, diphthongs, nasals, etc. Other than compound or foreign words, Thai words are usually monosyllabic. Most of the syllables are associated with tonal information, and there are 5 types of tones: mid, low, falling, high and rising. Inspired by Mandarin pronunciation dictionary [1] which also includes tonal information, the vowels and diphthongs in Thai are further classified according to the tone classes of their syllables. In such case, Thai dictionary consists of 153 tonal phonemes.

### B. Text Data Collection

One of the challenges in text data collection is to collect a large amount of data from different domains as the ASR systems aim for general purpose applications. We follow the approach in [17-18] to crawl the Web to collect large text corpora.

For each language, we choose a list of 10,000 most frequent words from its training transcription, and generate over tens of thousands of queries by randomly choosing any two words in the list. We send the queries to a number of commercial search engines and collect the URLs returned for each query. For the text collection of Bahasa Indonesia, we limit the search engines to the returned web pages from the .id domain, which is to avoid the returned web pages of other languages (e.g. Bahasa Malaysia). Pairs of words are used because there are fewer web pages in these two languages and using more than two words as a query usually cannot return enough URLs. To ensure the coverage of less frequent words in our collected corpora, we also use each of the remaining (or less frequent) words as a single search query to collect URLs. Duplicate URLs and non-HTML URLs (e.g. MS PowerPoint, JPG and PDF as extension) are removed, and the remaining HTML pages are downloaded for further data cleansing. The main text is extracted from each HTML page by detecting and removing the HTML tags, JavaScript and other non-linguistic materials.

## IV. EXPERIMENTS

### A. Data

By following the guidelines proposed in Section II, 800 hour speech data was collected from mobile systems for each language. The native speakers were recruited from the whole countries so the corpora covered as many accents as possible. Table I shows statistical information of the 2 corpora. About 13,000 utterances from each corpus were retained for testing purpose.

By applying the proposed approaches in Section III, pronunciation dictionaries were generated and the numbers of word entries are shown in Table I. The web page data was also collected from the Internet.

TABLE I. STATISTICS OF SPEECH CORPORA

|  | Bahasa Indonesia | Thai |
|---|---|---|
| Total No. of Speakers | 899 | 800 |
| Total No. of Utterances | ~1.1 million | ~1.0 million |
| Total Durations | 800 hours | 800 hours |
| Pronunciation Dictionary Sizes | ~111,000 | ~223,000 |

### B. Experimental Setup

The acoustic feature consists of 13 dimensional MFCC feature, 1 dimensional tone feature, and their $\Delta$, $\Delta\Delta$ and $\Delta\Delta\Delta$, resulting in 56 dimensions. The acoustic model is a HMM-DNN hybrid model.

The HMM acoustic model contains 35 monophones for Bahasa Indonesia and 153 monophones for Thai, respectively. In both the models, the context-dependent triphones are modeled by about 8,500 tied states. The final model was trained with DNN [1] on top of the model trained using maximum mutual information (MMI) technique. The DNN model has 5 hidden layers and each layer has 1024 nodes. The state-level minimum Bayes risk (sMBR) [19] was also applied to further refine the DNN model. The Kaldi toolkit was used to build the AM.

To evaluate the AM performance, a 3-gram web LM was trained for each language using the text data collected in Section III.

### C. Experimental Results

Table II shows the experimental results of both ASR systems in terms of WER.

The DNN-sMBR model of Bahasa Indonesia could achieve 13.6% WER. Such performance is usually better than the systems of many other languages according to our experience. It is close to the performance of our Malay ASR system. It is reasonable because Bahasa Indonesia and Malay are from the same language family and they are quite similar to each other. Both of them have more than 30 monophones which are less than many other languages so there is less confusion among phonemes. This probably makes the words of these two languages be recognized relatively easier.

TABLE II. WORD ERROR RATES USING 3-GRAM WEB LM

|  | Bahasa Indonesia | Thai |
|---|---|---|
| MMI | 20.8 | 42.8 |
| DNN | 15.1 | 38.8 |
| DNN-sMBR | 13.6 | 37.6 |

The DNN-sMBR model of Thai achieved 37.6% WER, and it is higher than many other languages. There might be a few reasons: Firstly, Thai is a tonal language and each vowel sound can be associated with 5 different tones. Hence, 5 different monophones should be created for a vowel sound. It increases the degree of confusion among all the monophones. Secondly, Thai words can be either mono-syllable or multi-syllable so that different segmentation methods may result in different WERs.

Thirdly, there could be domain mismatch between the test data and the web LM.

To further verify the Thai AM, extra work was carried out on the LM to reduce the effect of domain mismatch. The web LM was linearly interpolated with the LM trained using the training data transcripts. With the interpolated LM, the WERs of different AMs were reduced by 23.1%-27.4% relatively, as shown in Table III.

TABLE III. WORD ERROR RATES OF THAI ASR USING INTERPOLATED LM

|  | *Thai* |
|---|---|
| MMI | 32.9 |
| DNN | 29.0 |
| DNN-sMBR | 27.3 |

## V. SERVER-CLIENT APPLICATION

The trained AMs and LMs have been included in our cloud-based ASR system. The server manages a pool of multilingual decoders. Each decoder is able to process multiple recognition requests simultaneously. The communication between client and server is through TCP/IP packets with the pre-defined protocol. Figure 2 illustrates a standard speech recognition process. An end user can speak through an audio capture device (e.g. smart phone). While the user is speaking, the chunks of speech signal are packaged according to the protocol and they are being streamed to the server. Based on the protocol, the server assigns the chunks of speech signal to the corresponding decoder for processing. The transcribed text is sent back to the end user.

For client side development, our ASR system provides software development kit (SDK) in different programming languages, e.g. objective-c for iOS devices and Java for Android devices. The stand-alone SDK can provide the ASR service of multiple languages.

## VI. CONCLUSIONS

This paper reported our effort in developing ASR systems for Southeast Asian languages. The strategies and guidelines of collecting scripted speech data and text data were presented. The collected resources of Bahasa Indonesia and Thai have been applied to build general-purpose ASR systems. In our future work, other deep learning techniques such as Bidirectional Long Short-Term Memory Networks and Time Delay Neural Networks will be used in our AMs.

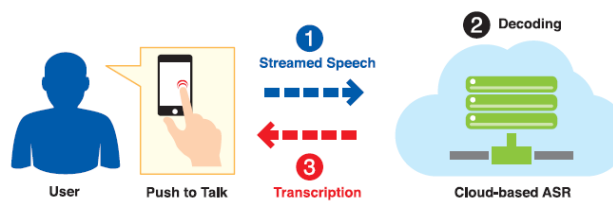

Figure 2. Illustration of Our Cloud-based Speech Recognition System